\documentclass{bmvc2k}

\usepackage{times}
\usepackage{soul}
\usepackage{url}
\usepackage{amsmath}
\usepackage{amsthm}
\usepackage{booktabs}
\usepackage{algorithm}
\usepackage{algorithmic}
\usepackage{wrapfig}
\usepackage{tikz}
\usepackage{comment}
\usepackage{color}
\definecolor{ch}{RGB}{12,15,205}
\usepackage{multirow}
\usepackage{adjustbox}
\usepackage{makecell}
\usepackage{amsfonts,amssymb}
\usepackage{subfigure}
\usepackage{graphicx}

\usepackage{float}

\newfloat{figtab}{htb}{fgtb}
\makeatletter
  \newcommand\figcaption{\def\@captype{figure}\caption}
  \newcommand\tabcaption{\def\@captype{table}\caption}
\makeatother

\usepackage[switch]{lineno}

\title{RBFormer: Improve Adversarial Robustness of Transformer by Robust Bias}

\addauthor{Hao Cheng}{ hcheng046@connect.hkust-gz.edu.cn}{1}
\addauthor{Jinhao Duan}{jd3734@drexel.edu}{2}
\addauthor{Hui Li}{hui01.li@samsung.com}{3}
\addauthor{Lyutianyang Zhang}{lyutiz@uw.edu}{4}
\addauthor{Jiahang Cao}{ jcao248@connect.hkust-gz.edu.cn}{1}
\addauthor{Ping Wang}{ping.fu@mail.xjtu.edu.cn}{5}
\addauthor{Jize Zhang}{cejize@ust.hk}{6}
\addauthor{Kaidi Xu}{kx46@drexel.edu}{2}
\addauthor{Renjing Xu}{renjingxu@ust.hk}{1}

\addinstitution{
 The Hong Kong University of Science and Technology (Guangzhou), Guangzhou, China
}
\addinstitution{
 Drexel University, Philadelphia, USA
}
\addinstitution{
 Samsung Research and Development Institute China Xi'an, Xi'an, China
}
\addinstitution{
 University of Washington, Seattle, USA
}
\addinstitution{
  Xi'an Jiaotong University, Xi'an, China
}
\addinstitution{
 The Hong Kong University of Science and Technology, Hong Kong SAR
}

\runninghead{Cheng et al.}{RBFormer}

\begin{document}

\maketitle

\begin{abstract}
Recently, there has been a surge of interest and attention in Transformer-based structures, such as Vision Transformer (ViT) and Vision Multilayer Perceptron (VMLP).
Compared with the previous convolution-based structures, the Transformer-based structure under investigation showcases a comparable or superior performance under its distinctive attention-based input token mixer strategy.
Introducing adversarial examples as a robustness consideration has had a profound and detrimental impact on the performance of well-established convolution-based structures. This inherent vulnerability to adversarial attacks has also been demonstrated in Transformer-based structures.
In this paper, our emphasis lies on investigating the intrinsic robustness of the structure rather than introducing novel defense measures against adversarial attacks.
To address the susceptibility to robustness issues, we employ a rational structure design approach to mitigate such vulnerabilities. Specifically, we enhance the adversarial robustness of the structure by increasing the proportion of high-frequency structural robust biases.
As a result, we introduce a novel structure called Robust Bias Transformer-based Structure (RBFormer) that shows robust superiority compared to several existing baseline structures. Through a series of extensive experiments, RBFormer outperforms the original structures by a significant margin, achieving an impressive improvement of $+16.12\%$ and $+5.04\%$ across different evaluation criteria on CIFAR-10 and ImageNet-1k, respectively.
\end{abstract}

\section{Introduction}
Convolutional Neural Networks (CNNs) have achieved breakthroughs in many domains~\cite{he2016deep,shi2020loss,yuan2020attribute,yuan2023remind,duan2023improve, cheng2022more}. However, adversarial examples~\cite{szegedy2013intriguing,xu2020adversarial,duan2023improve, ye2019adversarial, cheng2022more} as the inherent vulnerability of model structures has been extensively observed in CNNs across diverse contexts.
In allusion to alleviating this vulnerability, adversarial training~\cite{madry2017towards} as the most successful robust boosting method has also been proposed.
In addition to traditional CNNs, ViT~\cite{dosovitskiy2020image} and its subsequent studies~\cite{liu2021swin,girdhar2019video,chen2021mvt,neimark2021video,arnab2021vivit,yang2020learning,benz2021adversarial}, which are inspired by the transformer-based architectures~\cite{dosovitskiy2020image}, have been assumed as a novel base structure for solving various computer vision tasks.
Following the continual research about the characteristics of Multi-head Self-Attention (MSA),
\cite{dong2021attention} has found that the overmuch usage of MSA might adversely influence the Transformer performance and lead the entire output to converge exponentially to a rank-1 matrix. However, the Skip-Connections and MLP sub-blocks could mitigate and avoid this rank collapse phenomenon. This finding demonstrates that the MSA is not the most essential factor for the success of the Transformer, but the structure itself is. This phenomenon could quickly transfer to ViT and VMLP \cite{tolstikhin2021mlp,yu2021metaformer,reiser2021kilonerf}, which illustrates the structure like ViT but removes attention blocks.
About these two new structure, there are various works~\cite{ying2021transformers,he2016deep,zagoruyko2016wide,simonyan2014very,raghu2021vision,shao2021adversarial,yu2021rethinking} put their first focus on analyzing the difference between Transformer-based and Convolution-based structures, which ViT and CNN typically represent.  
They indicate that CNN holds the inductive bias compared with ViT. The core concept of inductive bias is the locality, which will make CNN focus more on low-level local information, but ViT pay more attention to high-level global information. Therefore, the convolution structure could be understood as a low-level or high-frequency structure. 
For robustness concerns, recently, \cite{naseer2021intriguing,mahmood2021robustness,shao2021adversarial,paul2021vision} claim that Transformer-based structures also exist adversarial vulnerability and could also be alleviated by adversarial training.
However, the current research in this field primarily focuses on studying basic robust features and enhancing the compatibility between existing defensive methods and established structures. To make a robust improvement based on the structural design level,
we may ask: 

\textit{Can we improve the robustness of original ViT/VMLP by rational structure design?}

\begin{figure}[h]
\centering
\begin{minipage}[t]{0.48\textwidth}
\centering
\includegraphics[width=6.5cm]{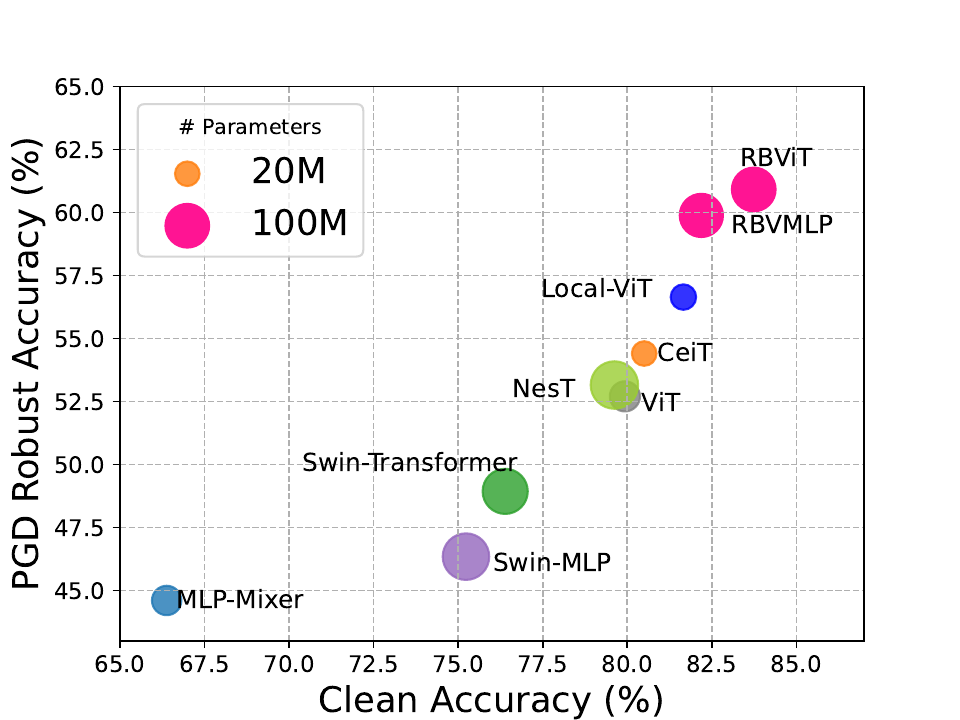}
\end{minipage}
\begin{minipage}[t]{0.48\textwidth}
\centering
\includegraphics[width=3.5cm]{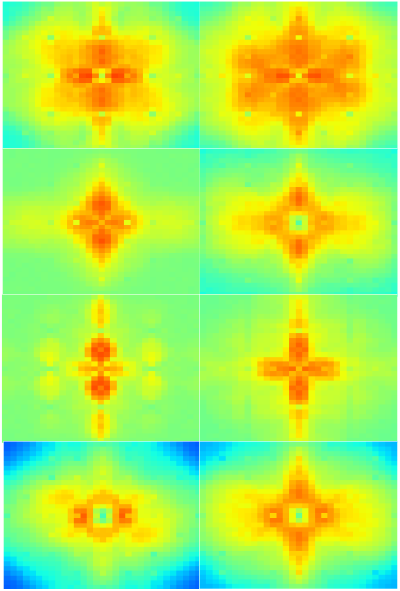}
\end{minipage}
\label{fig:base/freq1}
\caption{\footnotesize{Left: Comparison Results of RBFormer 
 (RBViT/RBVMLP) with current SOTA in clean/robust accuracy and model size. Right: Fourier Heatmap, from top to bottom, ViT/VMLP (internal Left/Right), adding convolution operation to embedding, embedding + block aggregation, and embedding + block aggregation + CMLP} }
\vspace{-3mm}
\end{figure}

\begin{figure*}[t!] 
\centering
\includegraphics[width=0.9\textwidth]{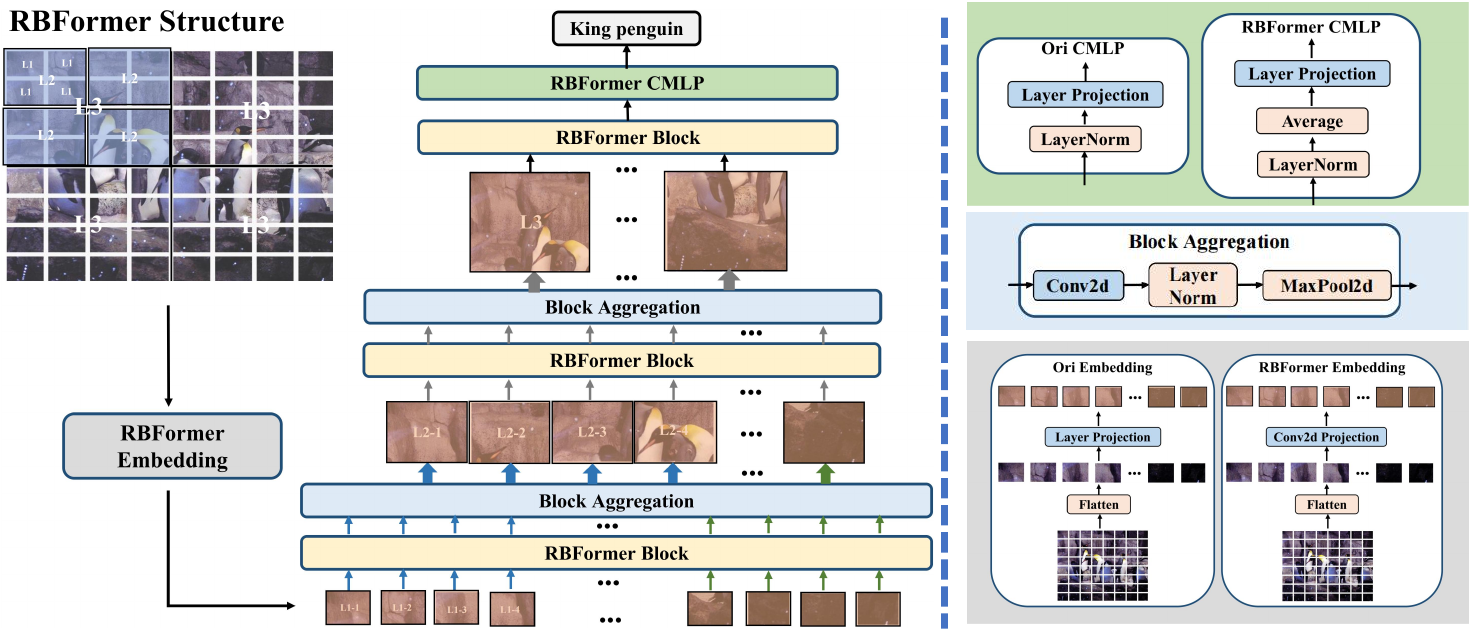}
\caption{\footnotesize{The RBFormer structure can be summarized as follows:
(a) The left sub-figure illustrates the general architecture of RBFormer.
(b) The right sub-figures depict the specific utilization of the CMLP block, block aggregation, and embedding, showcasing the modifications compared to the original structures
}} 
\label{Fig:main} 
\vspace{-5mm}
\end{figure*}

To pursue the answer to this question, the Transformer-based structures could be divided into three essential components: 1) Embedding, 2) Token-Mixer (TM) block; 3) Classifying MLP (CMLP) block,  and two training facilitation techniques: 1) Normalization, 2) Skip-connection. The TM block could be further divided into MSA sub-block and MLP sub-block.
Based on prior research, ViT/VMLP models differ from CNNs in their focus on high-level, low-frequency information, leading to a more global representation of images. While CNNs rely on an inductive bias that emphasizes locality, ViT/VMLP models prioritize capturing global context, allowing them to incorporate high-level, low-frequency details directly. This distinction highlights how the inductive bias in CNNs makes them more sensitive to high-level, low-frequency information.
In the meantime, \cite{yin2019fourier} analyzes the adversarial training from the Fourier perspective.
Therefore, introducing the high-frequency structure is a feasible way to improve adversarial robustness, and it could be called robust bias. Currently, there are two ways of strengthening the inductive bias or locality in ViT/VMLP, except for directly adding convolution operation to structural components~\cite{xu2021vitae, wu2021cvt, li2021localvit}, introducing multi-hierarchy layer stacking strategies~\cite{liu2021swin, zhang2021aggregating, wu2021cvt}, which could boost local or low-level visual structure by amplifying the ability of cross-patch information communication, is another good way. 
Following these two ways of increasing robust biases to the original structure, we could finally obtain RBFormer with better robustness after the comprehensive evaluation. The evaluation includes employing PGD~\cite{madry2017towards}, Auto-Attack~\cite{croce2020reliable}, frequency heat map~\cite{yin2019fourier}, and local Lipschitz~\cite{huang2021exploring} techniques to assess the clean and robust performance of various structure designs with different robust biases under natural and robust training scenarios. Eventually, we can provide insightful answers to the initial question and offer the following contributions:

\begin{itemize}
    \item After meticulous analysis and experiments, we demonstrate the effect of two distinct robust biases toward the robustness of Transformer-based structural designs.
    \item Evaluating various experimental findings leads to a deeper comprehension of robust biases integration characteristics, thereby informing the design of Transformer-based structures. Ultimately, we propose the RBFormer  (RBViT/RBVMLP), which exhibits the most robust performance, as depicted in Fig.~\ref{Fig:main}.
    \item According to the comparison results depicted in Fig.~\ref{fig:base/freq1}, the RBFormer structure outperforms recent popular adopted structures in terms of robustness. Specifically, RBFormer exceeds the original structures by a significant margin of $+16.12\%$ and $+5.04\%$ under various evaluation methods in CIFAR-10 and ImageNet-1k.
\end{itemize}


\section{Exploring Logic and Experimental Setup}
As mentioned earlier, this section aims to establish a coherent, logical flow to address the question while outlining the general experimental setup.
Firstly, designing Transformer-based structures according to the robust consideration is explored. Then, the original ViT or VMLP structures and our particular implementation are briefly introduced. Finally, we identify convolution operation and multi-hierarchy layer stacking strategy as our two potential available robust biases and further delve into their specific introducing methods.

\paragraph{Robust Consideration:}
The previous research work~\cite{yin2019fourier} on the relationship between structural robustness and frequency analysis first inspires us. Through the theoretical analysis of the activation and loss function for the most straightforward classification task and further visual experimental verification, the high-frequency structure could be named the robust bias to facilitate the robustness of Transformer-based models.

Previous works~\cite{xie2020smooth,gowal2019scalable} mention that generating more complicated or worse adversarial examples in the inner maximization would be beneficial to promote the final robust performance. About how to improve this inner optimization, \cite{yin2019fourier} explores the robust characteristic from the frequency domain and claims that adversarial training is a process for moving more structured focuses to high-frequency information. Consequently, the target of finding more challenging adversaries is equivalent to facilitating the ability to concentrate features on high-frequency domains.  

Adversarial training~\cite{madry2017towards} as a min-max optimization is mainly pursuing adversarial perturbation $\delta$ within $\ell_p$-ball constraint $\boldsymbol{\Delta}_p$ in the inner maximization process.
Inverse Discrete Fourier Transform (IDFT) is $x = \textit{IDFT}[X] = \frac{1}{N}\sum\nolimits_{k}XW_N^{-kn}, \, k = 0, ..., N-1$, where $W=e^{-j\frac{2\pi}{N}}$, $x$ and $X$ is the input in the temporal and frequency domain, $N$ is the transform interval length, all of the parameters are positive values. And IDFT is monotonically increasing. Fig.~\ref{fig:math} is the activation function (Sigmoid) and loss function (Cross-entropy loss) for the most straightforward two-class classification problem. Since the Sigmod function is also monotonically increasing, when the input of Sigmod moves to frequency values from $X_1$ to $X_2$, the output will also change from $y_1$ to $y_2$. In the cross-entropy loss of two labels, the possible value range of loss would also extend from $l_1$ to $l_2$. Consequentially, when the input frequency $X$ is more toward the high-value region, this simple classification task will result in a broader range of possible loss values like $l_1$ to $l_2$. In a word, higher-frequency information exploration will lead to more intricate adversarial examples since the inner max in adversarial training is pursuing higher loss value within $\ell_p$-ball constraint.

Recently, there are various works~\cite{he2016deep,zagoruyko2016wide, raghu2021vision,shao2021adversarial} certify that convolution operation could be used to promote the extracting ability of high-frequency information. Thus, we modify the embedding, block aggregation, and CMLP to a new version with convolution operation presented in Fig.~\ref{Fig:main}. This modification could also be illustrated through straightforward tools Fourier Heatmap~\cite{yin2019fourier} for the last feature map. 
The focus of the structure on high-frequency information would increase when the high-light concentration moves closer to the center.
The right subfigure in Fig.~\ref{fig:base/freq1} is the Fourier heatmap for the original ViT/VMLP (Left Column/Right Column), and the original ViT/VMLP is inserted convolution operation from top to bottom. The structures will focus more on the center or high-frequency domain when improving the proportion of convolution operation. And the following results in Section 3 and the Appendix will also indicate their better robustness. The convolution operation or high-frequency visual structure could be named the robust bias here. Additionally, we term an assumption: \emph{If we continually improve the proportion of this robust bias, would the whole model be more robust?} To validate this assumption, we would figure out which types of robust bias we can choose and how to design the corresponding experiments in the following sections.

\textbf{Structural Components:}
According to the previous introduction of RBFormer, all structures could be preliminarily separated into three main components: Embedding, TM block, and CMLP block. The TM block includes the MSA and MLP sub-block. This sub-section will present the particular structural composition of ViT/VMLP and our implementation. The detailed theoretical calculation is in Appendix~\ref{math2}.

\textbf{Original Components of ViT/VMLP:} 
(1) Embedding:
The embedding function is to transform the original images into the embedding tokens.
Embedding could be divided into two steps: Step 1 is to execute dimension transform in Eq.~\ref{eqn:ViTstep1}, and Step 2 in Eq.~\ref{eqn:ViTstep2} will add a learnable 1D positional embedding $E_{pos}$, which contains the positional information under the patch segment phase, to the token vector.
Additionally, similar to BERT~\cite{devlin2018bert}, ViT also adopts [class] token (CLS) to do classification.
(2) TM Block:
TM Block mixes the embedding tokens to capture the inner features of input images and mainly consists of two sub-blocks. For ViT, the first is the particular MSA sub-block, and the second is the MLP sub-block, which is constituted by two projecting layers with a GELU non-linearity. And VMLP, such as Mixer-MLP~\cite{tolstikhin2021mlp} and PoolFormer~\cite{yu2021metaformer}, would have two MLP sub-blocks since the original MSA sub-block would be replaced by MLP sub-block
Additionally, two crucial training facilitation techniques, LayerNorm (LN) and Skip-connection or Residual (Res), are adopted in both phases here.
(3) CMLP Block:
CMLP block is the final main component that constitutes two MLP sub-blocks with a GELU non-linearity. This block would also include LN and Res.

\textbf{Our Implementation}: 
(1) Embedding: Recent experimental findings~\cite{mao2021towards,wu2021cvt,zhang2021aggregating,chen2021mvt} indicate that removing the CLS token, which is used in the original ViT, can potentially enhance the performance. This kind of removal eliminates unnecessary components and reduces redundant computational costs.
Consequently, we directly average (AVG) all token vectors as the input of the TM block. 
(2) TM Block:
In the TM block, the general structure and components of the MSA and MLP sub-blocks will be retained, preserving similarity with the original ViT/VMLP models.
However, a modification was made to the Linear Layer within these two sub-blocks. It will be replaced with either Conv2d or Conv1d, depending on the insertion of the convolution operation, to ensure proper dimensional transformation.
About LN and Res, \cite{yin2019fourier} mentions that the normalization operation would play a significant role in analyzing adversarial robustness in the frequency domain. Therefore,  we will hold the LN as another main research object and put the Res as the ablation study in the Appendix.
(3) CMLP Block:
As presented in Fig.~\ref{Fig:main}. Since the CLS is removed in embedding, CMLP comprises an AVG pooling layer to do the average of every patch.

\textbf{Robust Biases Introduction:}
In this subsection, two potentially available robust biases are declared. One is the convolution operation explained above. The multi-scale hierarchy layer stacking strategy is another implicit robust bias that can facilitate the capturing ability of high-frequency information.

\textbf{Convolution Operation}
as a normal replacement of the projection layer used in the ViT/VMLP could be easily introduced into the following components.  
For embedding, there are two ways of executing it. The first is Convolution Embedding (CONV), which inserts some convolution projection structure before the original embedding. The second Projection Convolution Embedding (PCONV) directly adopts the convolution map to transform the dimension. 
The MSA sub-block, MLP sub-block, and CMLP block could directly adopt the convolution layer to replace the normal feed forward projection.

\textbf{Multi-hierarchy Layer Stacking:}
Four layer stacking strategies inspired by some recent works~\cite{xu2021vitae, liu2021swin, zhang2021aggregating, li2021localvit} are explored here and could incorporate convolution operations by using different inserting methods. We introduce them as follows:

\uppercase\expandafter{\romannumeral1}. \textbf{The original ViT Structure (OriViT)} directly utilizes the initial ViT structure~\cite{dosovitskiy2020image} as the layer stacking strategy. Since the embedding of OriViT is fixed and outputs a 1D token, the PCONV embedding and CONV TM block are unsuitable for OriViT.

\uppercase\expandafter{\romannumeral2}. \textbf{CNN-based Structure} would introduce the resolution and channel change process in the multi-hierarchy structure design by imitating the dimension change of CNN~\cite{wu2021cvt,li2021localvit,chen2021mvt,arnab2021vivit}.
Concretely, the core of this strategy is to directly introduce 2D images as input and keep the 2D dimension in the inner processing step. PCONV embedding, Conv TM block, and CONV MLP block could all be comprehended in this stacking strategy.

\uppercase\expandafter{\romannumeral3}. \textbf{Swin-based Structure~\cite{liu2021swin} (Swin)} achieves SOTA performance in various computer vision tasks. Swin modifies the original MSA sub-block to Window-based (WB) MSA and Shifted Window-based (SWB) MSA. WBM is trying to split each patch into a smaller sub-patch further. SWBM would introduce connections across windows and reinforce the ability to use local information. 
In a word, WB MSA and SWB MSA do not change the essential components of MSA but introduce additional patch splitting and window shift operation.
Since the essential components of Swin are just like OriViT except for the TM block, we could introduce convolution operation to each element of it directly. Swin-based VMLP~\cite{liu2021swin} has similar components and convolution operation introducing strategy as Swin-based ViT.

\uppercase\expandafter{\romannumeral4}. \textbf{Image Pyramid Structure (ImagePy)} as another stacking strategy that is inspired by NesT~\cite{zhang2021aggregating}. 
ImagePy first splits and then aggregates non-overlap image patches in a hierarchy way, and it does not need the cooperation of any component modification. 
In a word, except for the dimensional transformation limitation of PCONV, ImagePy is very flexible, and the convolution operation could be introduced to components in any way.

\section{Experiments and Analysis}
In the experiments, we adopt two popular datasets (i.e., CIFAR-10 and ImageNet-1k) to explore more robust structures by gradually increasing the proportion of robust bias.
For the specific evaluation, we use $\ell_{\infty}$-PGD~\cite{madry2017towards} and Auto-Attack~\cite{croce2020reliable}, which is an ensemble of white-box and black-box attacks, as the robust validating metrics. 
In CIFAR-10, we use three $\epsilon$ values for evaluating naturally trained models: $1/255$, $2/255$, and $3/255$. 
For adversarially trained models, $8/255$ as the most adopted worst value is selected.
For ImageNet-1k, we mainly evaluate the structural performance under the robust cases and use $4/255$ as default $\epsilon$. 
All models presented in this paper are trained from scratch without any transfer learning strategy.
Additionally, when doing the detailed analysis in CIFAR-10, two popular tools (Fourier heatmap~\cite{yin2019fourier} and Local Lipschitz~\cite{huang2021exploring}) are utilized to help understand the specific impacts of adding robust bias.
Fourier heatmap could measure the sensitivity of models when encountering noises in diverse frequency domains.
Each pixel of the heatmap is scaled to $[0, 1]$ and refers to the error rate after adding frequency noise with the pixel's coordinate as a basis for natural examples. For the Local-Lipschitz constant, as a numerical evaluation index of robustness, the lower value of a structure indicates its smoother and more robust characteristic.

This section is organized as follows: Section 3.1: Experimental setup of adding robust bias structure; Section 3.2: Validate and analyze the particular results and characteristics when inserting different robust bias visual structures to corresponding components.
Section 3.3: Compare our RBFormer (RBViT/RBVMLP) with various popular baseline structures.

\subsection{Structures with Various Robust Biases }
This subsection indicates how to incorporate two robust biases: (1) convolution operation and (2) multi-hierarchy layer stacking strategy to the original ViT/VMLP and finally propose RBFormer (RBViT/RBVMLP).
Specifically, for adding convolution operation, according to the explanation of Section 2.1, there would be three components, (1) Embedding, (2) TM Block, including MSA and MLP sub-block, (3) CMLP block, and one technique, (4) LN, that could be inserted. About the multi-hierarchy layer strategy, (1) OriViT~\cite{dosovitskiy2020image}; (2) CNN-based structure~\cite{wu2021cvt,d2021convit,xu2021vitae}; (3) Swin~\cite{liu2021swin} and (4) ImagePy~\cite{zhang2021aggregating} would be our research objects. In the specific implementation process, the above-introduced robust biases are not all compatible with each other because of the dimension matching. The concrete component combinations are as follows:

\textbf{Convolution Operation:}
To explore the robust influence of inserting convolution operation on each component and the presence or absence of LN, we modify each component according to the presence of convolution-adding degree and LN in Table~\ref{tab:cifar}. Additionally, the multi-hierarchy layer stacking strategy, as a structure that can bring that similar utility as the convolution operation, is also included in Table~\ref{tab:cifar}.
For the specific options: 
There would be two options in embedding: Original (Ori) and CONV embedding. Since the dimension mismatch issues between CONV TM block/PCONV with the OriViT, we will not consider the CONV TM block and PCONV here. Additionally,  accompanied by whether to add the convolution operation, CMLP  will have two choices here: Original (Ori) MLP, Convolution (CONV) MLP. Finally, 
Norm also has two options here: Layernorm (LN) or None.
After permutation and combination among all possible choices, the robust performance of 
Eight structures (a) to (h) are analyzed in Table~\ref{tab:cifar}, Fig.~\ref{fig:imageresults}, and Appendix~\ref{exdetails}

\begin{table}
\small
\centering
\resizebox{\linewidth}{!}{ 
\begin{tabular}{cccccc|c|c|c|c}
\toprule
\multicolumn{6}{c|}{ViT/VMLP}                                                                      & \multirow{2}{*}{\begin{tabular}[c]{@{}c@{}}Clean\\ Accuracy\end{tabular}} & \multirow{2}{*}{\begin{tabular}[c]{@{}c@{}}PGD\\ (8/255)\end{tabular}} & \multirow{2}{*}{\begin{tabular}[c]{@{}c@{}}Auto-\\ Attack\\ (8/255)\end{tabular}} & \multirow{2}{*}{\begin{tabular}[c]{@{}c@{}}Lipschitz\\ Constant\end{tabular}} \\ \cline{1-6}
\multicolumn{1}{c|}{\begin{tabular}[c]{@{}c@{}}Components\\ Combine\end{tabular}} & \multicolumn{1}{c|}{Embedding} & \multicolumn{1}{c|}{TM}     & \multicolumn{1}{c|}{CMLP}  & \multicolumn{1}{c|}{Norm} & \begin{tabular}[c]{@{}c@{}}Stacking\\ Structure\end{tabular} &                                                                           &                                                                        &                                                                         &                                                                               \\ \midrule
\multicolumn{1}{c|}{(a)}                                                          & \multicolumn{1}{c|}{Ori}       & \multicolumn{1}{c|}{Ori}    & \multicolumn{1}{c|}{Ori}  & \multicolumn{1}{c|}{None} & oriViT                                                       & 79.88/71.06                                                               & 52.66/45.56                                                            & 51.12/44.37                                                             & 159.2/163.2                                                                   \\
\multicolumn{1}{c|}{(b)-Ori}                                                          & \multicolumn{1}{c|}{-}         & \multicolumn{1}{c|}{-}      & \multicolumn{1}{c|}{-}    & \multicolumn{1}{c|}{LN}   & -                                                            & 79.93/66.38                                                               & 52.70/44.06                                                            & 51.45/43.10                                                             & 157.7/164.7                                                                   \\
\multicolumn{1}{c|}{(c)}                                                          & \multicolumn{1}{c|}{-}         & \multicolumn{1}{c|}{-}      & \multicolumn{1}{c|}{CONV} & \multicolumn{1}{c|}{None} & -                                                            & 82.81/78.83                                                               & 54.79/54.24                                                            & 53.83/53.69                                                             & 151.3/152.3                                                                   \\
\multicolumn{1}{c|}{(d)}                                                          & \multicolumn{1}{c|}{-}         & \multicolumn{1}{c|}{-}      & \multicolumn{1}{c|}{-}    & \multicolumn{1}{c|}{LN}   & -                                                            & 81.66/77.58                                                               & 54.69/51.00                                                            & 53.85/50.88                                                             & 152.7/157.5                                                                   \\
\multicolumn{1}{c|}{(e)}                                                          & \multicolumn{1}{c|}{CONV}      & \multicolumn{1}{c|}{-}      & \multicolumn{1}{c|}{Ori}  & \multicolumn{1}{c|}{None} & -                                                            & 82.77/77.86                                                               & 55.85/53.22                                                            & 54.98/52.89                                                             & 151.4/155.8                                                                   \\
\multicolumn{1}{c|}{(f)}                                                          & \multicolumn{1}{c|}{-}         & \multicolumn{1}{c|}{-}      & \multicolumn{1}{c|}{-}    & \multicolumn{1}{c|}{LN}   & -                                                            & 80.50/75.92                                                               & 54.40/50.89                                                            & 53.69/48.99                                                             & 153.1/162.3                                                                   \\
\multicolumn{1}{c|}{(g)}                                                          & \multicolumn{1}{c|}{-}         & \multicolumn{1}{c|}{-}      & \multicolumn{1}{c|}{CONV} & \multicolumn{1}{c|}{None} & -                                                            & 80.57/79.25                                                               & 55.63/53.81                                                            & 54.23/52.45                                                             & 146.3/148.5                                                                   \\
\multicolumn{1}{c|}{(h)}                                                          & \multicolumn{1}{c|}{-}         & \multicolumn{1}{c|}{-}      & \multicolumn{1}{c|}{-}    & \multicolumn{1}{c|}{LN}   & -                                                            & \textbf{82.35/81.42}                                                      & \textbf{56.41/56.89}                                                   & \textbf{56.12/57.02}                                                    & \textbf{140.3/141.3}                                                          \\ \midrule
\multicolumn{1}{c|}{(i)-CVT}                                                          & \multicolumn{1}{c|}{PCONV}     & \multicolumn{1}{c|}{CONV}   & \multicolumn{1}{c|}{Ori}  & \multicolumn{1}{c|}{-}    & CNN-based                                                    & 79.62/77.62                                                               & 53.15/52.12                                                            & 52.11/50.21                                                             & 143.2/146.9                                                                  \\
\multicolumn{1}{c|}{(j)}                                                          & \multicolumn{1}{c|}{-}         & \multicolumn{1}{c|}{-}      & \multicolumn{1}{c|}{CONV} & \multicolumn{1}{c|}{-}    & -                                                            & 80.98/79.64                                                               & 57.67/56.83                                                            & 57.34/57.06                                                             & 136.9/138.1                                                                   \\ \midrule
\multicolumn{1}{c|}{(k)-Swin}                                                          & \multicolumn{1}{c|}{Ori}       & \multicolumn{1}{c|}{WBM+SWBM} & \multicolumn{1}{c|}{Ori}  & \multicolumn{1}{c|}{-}    & Swin-based                                                   & 76.39/75.23                                                               & 48.93/46.34                                                            & 47.64/45.21                                                             & 152.0/154.2                                                                   \\
\multicolumn{1}{c|}{(l)}                                                          & \multicolumn{1}{c|}{PCONV}     & \multicolumn{1}{c|}{-}      & \multicolumn{1}{c|}{CONV} & \multicolumn{1}{c|}{-}    & -                                                            & 80.08/78.34                                                               & 52.10/50.92                                                            & 50.48/50.22                                                             & 146.3/145.2                                                                   \\ \midrule
\multicolumn{1}{c|}{(m)-NT}                                                          & \multicolumn{1}{c|}{Ori}       & \multicolumn{1}{c|}{Ori}    & \multicolumn{1}{c|}{Ori}  & \multicolumn{1}{c|}{-}    & ImagePy                                                      & 76.22/75.94                                                               & 52.45/51.82                                                            & 51.28/49.14                                                             & 158.2/167.9                                                                   \\
\multicolumn{1}{c|}{(n)-RB}                                                          & \multicolumn{1}{c|}{PCONV}     & \multicolumn{1}{c|}{CONV}   & \multicolumn{1}{c|}{CONV} & \multicolumn{1}{c|}{-}    & -                                                            & \textbf{83.74/82.19}                                                      & \textbf{60.91/59.88}                                                   & \textbf{59.69/59.22}                                                    & \textbf{89.1/98.7}                                                            \\ \bottomrule
\end{tabular}
}
\caption{\footnotesize{The performance of our exploring structures under adversarial training in CIFAR-10.
All results are shown in \textbf{ViT/VMLP accuracy  (\%)}. This Table includes two robust bias explorations,
1) Structure (a)-(h): the exploration of convolution operation; 2) Structure (h)-(n): the exploration of multi-hierarchy layer stacking strategy. Among (a)-(n),  
some of them are corresponding to some current typical structures, (b) is the original ViT/VMLP~\cite{dosovitskiy2020image,tolstikhin2021mlp} (Ori), (i) is corresponding to CVT~\cite{wu2021cvt}/CVT-based VMLP (CVT) or CNN-based structures, (k) is Swin ViT/MLP~\cite{liu2021swin} (Swin),  (m) is the NesT~\cite{zhang2021aggregating}/NesT-based VMLP (NT). (n) is our final RBViT/RBMLP (RB).
}}
\label{tab:cifar}
\vspace{-5mm}
\end{table}

\textbf{Multi-hierarchy Layer Stacking Robust Bias:}
After discovering the best structure for introducing convolution operation robust bias, we further focus on multi-hierarchy layer stacking strategies. The robust performance is presented in Table~\ref{tab:cifar}, Fig.~\ref{fig:imageresults}, and Appendix~\ref{exdetails}. 
\uppercase\expandafter{\romannumeral1}. \textbf{OriViT:} 
We adopt this structure as our basic structure to explore the influence of adding convolution operation in the structure (a) to (h).
\uppercase\expandafter{\romannumeral2}. \textbf{CNN-based Structure:} 
For the CNN-based structure, its embedding and TM block should be fixed as PCONV embedding and CONV TM block. For the CMLP block, the original form could be replaced with CONV CMLP to increase its robust bias. Structures (i) and (j) are the evaluation of CNN-based structures.
\uppercase\expandafter{\romannumeral3}. \textbf{Swin:} 
The main distinction of Swin is to introduce cyclic shift operation by switching the original TM block to the WB and SWB token-mixer. In this case, the TM block could not be replaced as a CONV block. However, the original embedding and CMLP block could be substituted as PCONV and CONV. The performance of Swin is presented by structures (k) and (l).
\uppercase\expandafter{\romannumeral4}. \textbf{ImagePy:}
ImagePy is a simple stacking strategy compared with CNN-based structure and Swin without modifying any component. 
Except for its embedding that is modified to PCONV because of the 2D image dimension, both the form of TM and CMLP block should be similar to OriViT.
More detailed information, like layer number selection, distribution of each hierarchy, and others, are all offered in Appendix~\ref{exdetails}.

\subsection{Results Analysis}

\begin{figure}[h]
\vspace{1.5mm}
\centering
\begin{minipage}[t]{0.48\textwidth}
\centering
\includegraphics[width=7cm]{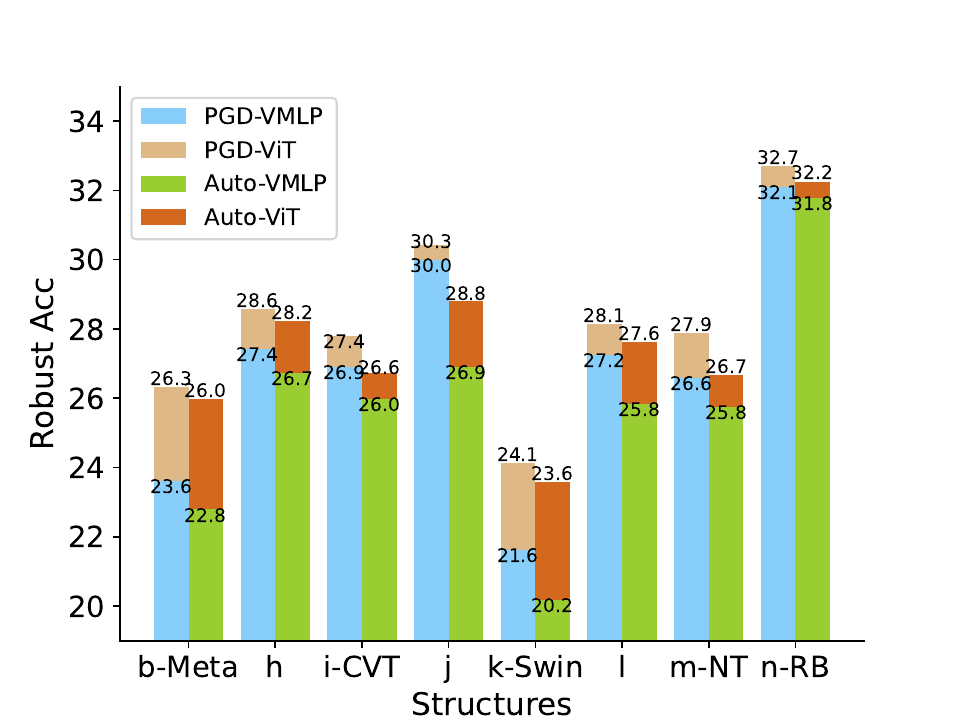}
\end{minipage}
\begin{minipage}[t]{0.48\textwidth}
\centering
\includegraphics[width=3cm]{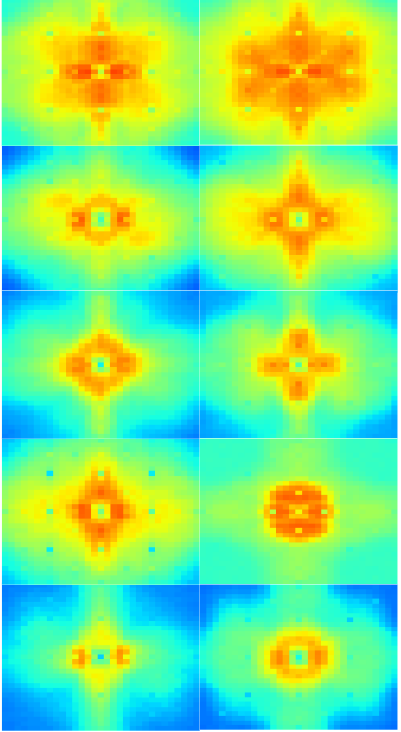}
\end{minipage}
\vspace{1.5mm}
\caption{\footnotesize{Left: ImageNet-1k results for ViT/VMLP structure.
PGD-VMLP/PGD-ViT and Auto-VMLP/Auto-ViT represent the robust accuracy under PGD and auto ($\epsilon=8/255$) attack. Right: Fourier heatmaps of Ori ViT/VMLP, some representative structures with convolution operation, and our RBFormer  (b, h, j, l, n) from top to bottom.}  }
\label{fig:imageresults}
\end{figure}

According to the adversarial robust performance of CIFAR-10 and ImageNet-1k in Table~\ref{tab:cifar}, Fig.~\ref{fig:imageresults} and Table~\ref{tab:imagenet}, we could conclude some interesting findings about the robust characteristic of transformer-based structures and propose our RBFormer:    

After comparing the performance of the (a)-(h) structure in Table~\ref{tab:cifar}, we obtain the conclusion that improving the proportion of convolution operation in ViT/VMLP-based models could availably boost the robustness of corresponding models. Undoubtedly, the presence of LN has less impact on robustness. Comparing the results among structure (a) to (h) in CIFAR-10, the robust results of ViT/VMLP under PGD attack and Auto-Attack with $\epsilon=4/255$ has at most 3.71\%/12.83\% and 4.67\%/12.92\% enhancement. And the Lipschitz-constant value comes into a 17.4/23.4 decrease (the lower, the better in robust case). Additionally, when ignoring the presence of LN, we adopt the Fourier heatmap on structures (b), (d), (f), and (h). As mentioned in Section 1.1, after adding more convolution operations from (b) to (h), the Fourier map concentrates more on the central zone, which indicates this model could capture more high-frequency information. The robust performance also increases from (a) to (h). Therefore, convolution operation as a high-frequency information-capturing structure or robust bias could promote robustness. In ImageNet-1k, the structure (a) and (h) in the left sub-figure in Fig.~\ref{fig:imageresults} could also conclude a similar observation.
After ensuring the positive effect of adding convolution operation, we target the 
oriViT, CNN-based, Swin, and ImagePy. The robust influence of these four strategies is our primary purpose in this subsection.
According to the PGD attack and Auto-Attack with $\epsilon=8/255$, as well as Lipschitz constant value from structures (h) to (n) in Table~\ref{tab:cifar}, Fig.~\ref{fig:imageresults} and Table~\ref{tab:imagenet}, we could acquire two observations about changing layer stacking strategy: (1) In each kind of layer stacking strategy, adding convolution operation to any components could generate a positive effect on improving robustness; (2) Not any layer stacking strategy could successfully introduce robust bias to boost robustness, like Swin in structures (k) and (l). The suitable layer stacking strategy should be a significant consideration when designing transformer-based structures. We also utilize the Fourier heatmap to analyze the characteristics of four strategies with the highest proportion of convolution operation. 
(structure h, j, l, and n). According to the experimental results in CIFAR-10, following the partial enhancement of clean accuracy, structure (n) could generate 8.21\%/15.82, 8.24\%/16.12\% robust accuracy improvement under PGD adversarial examples and Auto-Attack, 27.0/30.2 decrease in Lipschitz constant, and concentrating more on the central zone in the Fourier heatmap as Fig.~\ref{fig:imageresults}. In ImageNet-1k, compared with the original ViT/VMLP (b), structure (n) could generate 4.13\%/4.66\%, 4.03\%/5.04\% improvement under PGD attack and Auto-Attack.
Therefore, structure (n) would be our target RBFormer (RBViT/RBVMLP).

\subsection{Comparison Results of RBFormer with SOTA Baselines}
In this section, we mainly focus on the robust enhancement of our RBViT/RBVMLP (n) with some popularly used ViT~\cite{dosovitskiy2020image}/Mixer-MLP~\cite{tolstikhin2021mlp} (b), CVT/CVT-based VMLP~\cite{wu2021cvt} (i), Swin ViT/VMLP~\cite{liu2021swin} (k), NesT/NesT-based VMLP~\cite{zhang2021aggregating} (m) in Table~\ref{tab:cifar}, Fig.~\ref{fig:imageresults} and Table~\ref{tab:imagenet}. 
We first use PGD and Auto-Attack to compare our RBViT/RBVMLP with the other ViT/VMLP-based structures. Among these four values, our RBViT/RBVMLP could achieve at most 16.12\%, 
9.01\%, 14.01\%, and 9.56\% improvement  for CIFAR-10. And for ImageNet-1k in Fig.~\ref{fig:imageresults} and Table~\ref{tab:imagenet}, RBViT/RBVMLP could attain at most 5.04\%, 5.04\%, 11.59\%, and 6.02 \% enhancement. Additionally, in CIFAR-10, we also further adopt Lipschitz Constant (Lower values mean better robustness) to evaluate the robustness after comparing RBViT/RBVMLP (n) with those four ViT/VMLP-based structures (b), (i), (k), and (m), it can earn at most 68.6, 54.1, 62.9, 69.2 value decrease that means better robustness.
We also adopt the left figure in Fig.~\ref{fig:base/freq1} to illustrate the superiority of our RBFormer compared with ViT, CeiT~\cite{yuan2021incorporating}, Local-ViT~\cite{li2021localvit}, Mixer-MLP, NesT, CVT, Swin Transformer and Swin MLP~\cite{liu2021swin}. Our RBFormer (RBViT/RBMLP) could obtain the best robust performance compared with other baseline models.
Additionally, in contrast to \cite{mao2022discrete, qin2022understanding} that claim to specialize in improving the general robustness of Transformer-based structures, Table~\ref{tab:bench} shows that RBFormer also maintains performance advantages in both clean and adversarial cases.

\begin{table}[h]
        \setlength{\belowcaptionskip}{0cm}
        \setlength{\abovecaptionskip}{-0.07cm}
        \centering
        \label{tab:res}
	\scalebox{0.9}{
\begin{tabular}{c|cccc}
\hline
\multirow{3}{*}{\begin{tabular}[c]{@{}c@{}}Metric\\ (\%)\end{tabular}} & \multicolumn{4}{c}{ViT/VMLP}                                                                                       \\ \cline{2-5} 
& \multicolumn{2}{c|}{CIFAR-10}                                       & \multicolumn{2}{c}{ImageNet-1k}              \\ \cline{2-5} 
& \multicolumn{1}{c|}{Clean Acc}   & \multicolumn{1}{c|}{Adv Acc}     & \multicolumn{1}{c|}{Clean Acc}   & Adv Acc   \\ \hline
Mao et al.~\cite{mao2022discrete}                                                                      & \multicolumn{1}{c|}{83.13/\textbf{83.44}}            & \multicolumn{1}{c|}{52.79/47.13}            & \multicolumn{1}{c|}{60.46/58.81}            &    24.23/21.67       \\
Qin et al.~\cite{qin2022understanding}                                                                      & \multicolumn{1}{c|}{82.76/81.17}            & \multicolumn{1}{c|}{53.61/45.65}            & \multicolumn{1}{c|}{61.14/59.43}            &    26.55/24.52        \\
RBFormer                                                                    & \multicolumn{1}{c|}{\textbf{83.74}/82.19} & \multicolumn{1}{c|}{\textbf{60.91/59.88}} & \multicolumn{1}{c|}{\textbf{61.59/60.27}} & \textbf{32.71/32.09} \\ \hline
\end{tabular}
  }
  \vspace{0.4cm}
  \caption{\footnotesize{Comparing RBFormer with benchmarks under clean and adversarial case in CIFAR-10 and ImageNet-1k}}
  \label{tab:bench}
\end{table}

\subsection{The Affinity for Sparse Algorithms}
\begin{figure}[h!] 
\centering
\includegraphics[width=0.95\textwidth]{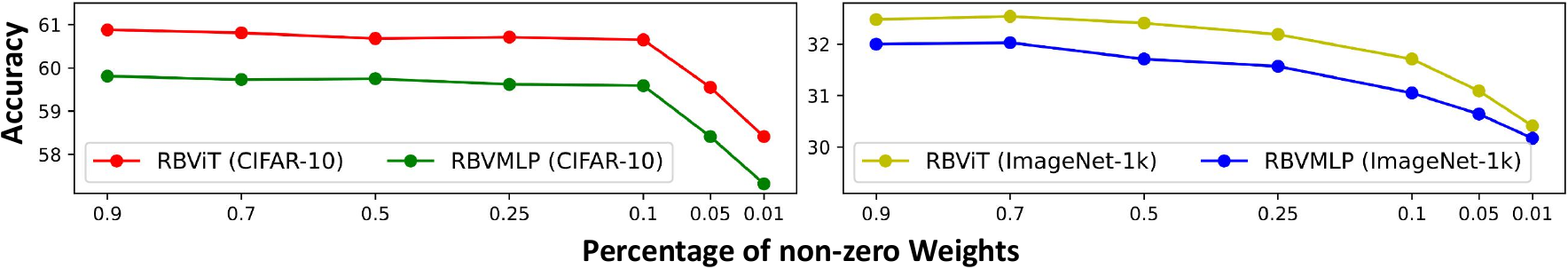}
\caption{\footnotesize{Applying irregular pruning to RBViT/RBVMLP in CIFAR-10 (Left) and ImageNet-1k (Right). }}
\label{fig:irr}
\end{figure}

RBFormer is realized through the process of rational component analysis and structure redesign, but instead, simply adding new parameters to increase the redundancy. 
This process involves only some modification in the dimensional transformation operation and does not significantly introduce computational complexity.
Additionally, RBFormer could adopt various model compression methods, such as model pruning, quantization, and sparse training, to further reduce our parameter numbers and model size without sacrificing robust performance. The most straightforward and least technically advantageous irregular magnitude pruning is adopted in Fig.~\ref{fig:irr}, the robust performance could be maintained under a low percentage of remaining non-zero weights.


\section{Conclusion}

In this paper, we perform an exhaustive examination of the critical components that notably influence the performance of Transformer-based structures. Our analysis systematically explores the impact of each component on robust vulnerability by experimenting with various combinations. Furthermore, we delve into a study of adversaries in the frequency domain, identifying robust biases that could potentially enhance adversarial robustness. Our proposed RBFormer integrates a carefully selected mix of these robust biases. Through rigorous experimental validation, we affirm that the RBFormer surpasses robust SOTA baselines, including currently prevalent structures and some methods that could be used to enhance general robustness. Therefore, introducing robust biases leads to a noticeable enhancement in overall performance, as evidenced by our study results.

\newpage

\bibliography{egbib}

\setcounter{section}{0}
\setcounter{figure}{0}
\setcounter{equation}{0}
\makeatletter 
\renewcommand{\thefigure}{A\@arabic\c@figure}
\makeatother
\setcounter{table}{0}
\renewcommand{\thetable}{A\arabic{table}}
\renewcommand{\theequation}{S\arabic{equation}}

\appendix
\onecolumn
\section*{Appendix}

\section{Experimental Details and Additional Restuls}
\label{exdetails}

\subsection{Experimental Detailed Setup}
About the particular layer number selection, referring to some current widely used multi-hierarchy structures Swin~\cite{liu2021swin}, NesT~\cite{zhang2021aggregating}, CVT~\cite{wu2021cvt} and etc~\cite{yuan2021incorporating,li2021localvit,yu2021metaformer}, we adopt 12 as our total number of layers. 
The simplest OriViT could directly set the layer number to be 12.
For a multi-hierarchy stacking structure, we adopt a three-stage hierarchy for CNN-based structures with $\{1,2,9\}$ for each stage. However, since the modification of Swin is on MSA and does not directly introduce convolution operation,  We would choose four steps with $\{2,2,6,2\}$ distribution for it.
For the ImagePy, the hierarchy distribution could be $\{2,2,8\}$ to maintain the properties of pyramid structures. 
About the training phase, in CIFAR-10, we respectively validate different structures under
natural and adversarial training cases. About ImageNet-1k, we mainly focus on exploring the robustness of adversarial training cases.
To generate adversarial examples in CIFAR-10, about natural case, we adopt the $\epsilon = 1/255, 
 2/255,  3/255$ with iteration = 10 and step size = 0.01 to attack the original evaluation model its robustness. In the adversarial case, we adopt the $\epsilon = 8/255$ with the same iteration and step size to generate adversarial examples and do training. For ImageNet-1k, adversarial examples are generated by using $\epsilon = 4/255 $ with iteration = $3$ and step size = $2\epsilon/3 $ referring to \cite{tang2021robustart}.
For the specific experimental results, Table~\ref{tab:imagenet} is the exact values of Fig.~\ref{fig:imageresults}. 
Additionally, according to \cite{dong2021attention} and further study above, the structural design of ViT is the most critical point to achieve better performance. The abuse of the attention block will make the overall structure collapse to the rank-1 matrix. For alleviating this collapse, skip connections are very crucial. MLP could help, but LN plays no role.
To verify the impact of this discovery on the design of robust transformer-based structures, we further explore the effect of Skip-Connection or Res as a facilitation technique towards robustness in the natural trained case as Table~\ref{tab:natural-alltrain} and adversarial training case as Table~\ref{tab:Adv-alltrainitem}.

\subsection{Experimental Analysis}
About the performance effect of skip-connection, MLP, and LN in Transformer\cite{dong2021attention}, the experimental results in 
Table.~\ref{tab:Adv-alltrainitem} and Table.~\ref{tab:natural-alltrain} show that the independent existence of skip-connection and MLP has a minor influence on the robustness. 
On the contrary, the effect of LN seems complex. When just removing LN, there is no performance change or even a little increase in some cases like (5)-(7), (9)-(11),  (13)-(15). After removing the skip-connection or MLP, further removing LN will make the structure not converge. For a more detailed statement, there will be the following changes about removing the LN layer:

\uppercase\expandafter{\romannumeral1}. According to the structures (1)-(3), (21)-(23), the performance of these component combinations without LN only has a minor drop compared with the original one;

\uppercase\expandafter{\romannumeral2}. According to structures (5)-(7), (9)-(11), (13)-(15), (17)-(19), some structures could have a better robust performance after removing the LN.

In a word, for the transformer-based structures in CIFAR-10, the existence of the Norm will not help the robustness, but removing it could even promote robustness a little in some cases. 
Furthermore, in ImageNet-1k, all structures would not converge after removing LN. Sequentially, we could acquire that the LN will 1) play no role or be little harmful to the robust performance of MetFormer and RBFormer structure with small training tasks (small model size or training datasets); 2) guarantee the training convergence under the large model size and extensive datasets. 
Consequently, based on the analysis above, we would still keep LN in our RBFormer to guarantee successful convergence in our training phase.

\section{Mathematical Analysis}
\label{math}

\subsection{ViT Structures Expression}
The mathematical representation of ViT structure: 
\begin{equation}\label{eqn:ViTstep1}
\begin{aligned}
        \textbf{X}_p = 
        &\mathcal{F}_{\text{DT}}(\textbf{X}) = [x^1_p; x^2_p;...;x^N_p] , 
        \\
    &\textbf{X} \in \mathbb{R}^{H\times W \times C},  \textbf{X}_p \in \mathbb{R}^{N \times (P^2 \cdot C)}
\end{aligned}
\end{equation}

\begin{equation}\label{eqn:ViTstep2}
\begin{aligned}
    \textbf{Z}_0
    =&\textbf{X}_p + \textbf{E}_{pos}=[x^1_pe; x^2_pe;...;x^N_pe] , \\  
    &\textbf{Z}_0 \in \mathbb{R}^{N \times (P^2 \cdot C)},  \textbf{E}_{pos} \in \mathbb{R}^{N \times (P^2 \cdot C)}
\end{aligned}
\end{equation}

\begin{equation}\label{eqn:ViTstep3}
\begin{aligned}
    \textbf{Z}^{'}_{l}=
    &Res(\mathcal{F}_{{\text{TM}}}(\mathcal{F}_{\text{LN}}(\textbf{Z}_{l-1}))) , \\  
    &l = 1, ..., L, \quad \textbf{Z}^{'}_{l} \in \mathbb{R}^{N \times (P^2 \cdot C)}
\end{aligned}
\end{equation}

\begin{equation}\label{eqn:ViTstep4}
\begin{aligned}
    \textbf{Z}_{l}=
    &Res(\mathcal{F}_{\text{MLP}}(\mathcal{F}_{\text{LN}}(\textbf{Z}^{'}_{l}))) , \\  &l = 1, ..., L, \quad \textbf{Z}_{l} \in \mathbb{R}^{N \times (P^2 \cdot C)}
\end{aligned}
\end{equation} 

\begin{equation}\label{eqn:ViTstep5}
    \textbf{y}=\mathcal{F}_{\text{CMLP}}(\mathcal{F}_{\text{AVG}}(Z_{L})) 
\end{equation}

\begin{figure}[h!]
\centering
\includegraphics[scale=0.5]{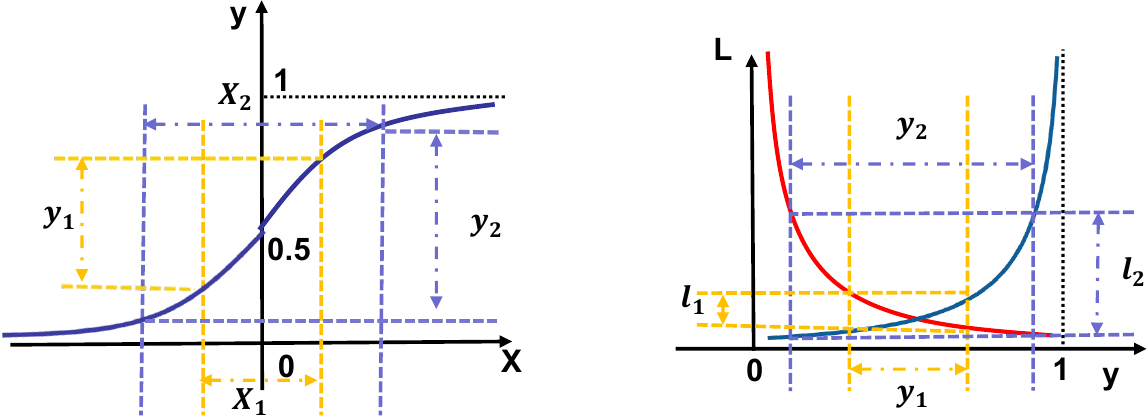}
\caption{\footnotesize{The activation function (left) and loss function (right) map.}}
\label{fig:math}
\vspace{-2mm}
\end{figure}

The embedding function transforms the original images into the embedding tokens. These generating tokens are corresponding processing objects of the following TM block.
Embedding could be divided into two steps. Step 1 is adopted to execute dimension transform in Eq.~\ref{eqn:ViTstep1}, which could reshape 2D image $x \in \mathbb{R}^{H\times W \times C}$ to a 1D sequence through flattening original 2D patches $x_p \in \mathbb{R}^{N \times (P^2 \cdot C)}$. $(H,W)$ is the resolution of the original image, $C$ is the channel number, $P$ is the patch size. And the number of patches will be $N=HW/P^2$.
After finishing Step 1, Step 2 in Eq.~\ref{eqn:ViTstep2} will add a learnable 1D positional embedding $E_{pos}\in \mathbb{R}^{N \times (P^2 \cdot C)}$ to the token vector. The introduction of this positional embedding is due to the loss of related positional information under the patch segmenting phase. 
Additionally, similar to BERT's~\cite{devlin2018bert} [class] token (CLS), ViT also adopts this component to do classification.
After generating embedding token $\textbf{Z}_0$, the next step of the transformer-based structure is to use TM Block to mix those embedding tokens and make the most significant efforts to capture the inner features. As we know before, two sub-blocks mainly constitute the TM Block. The first one is the particular \textit{MSA sub-block} (Eq.~\ref{eqn:ViTstep3}), and the second is the \textit{MLP sub-block} constituted by two projecting layers with a GELU non-linearity. For the VMLP structures, like the Mixer-MLP~\cite{tolstikhin2021mlp} and PoolFormer~\cite{yu2021metaformer}, they adopt \textit{MLP sub-block} to replace the initial \textit{MSA sub-block} and thus have two \textit{MLP sub-blocks}.
Apart from these two main constituting sub-blocks, Layernorm (LN) and Skip-connection, or Residual (Res), are also adopted in both phases as presented in  Eq.~\ref{eqn:ViTstep3}
 and \ref{eqn:ViTstep4}.
CMLP block is the final main component in Eq.~\ref{eqn:ViTstep5} and constitutes two MLP sub-blocks with a GELU non-linearity. 

\subsection{The Detailed Analysis of Robust Consideration}
\label{math2}
According to the equation Eq.~\ref{eqn:ViTstep1} to Eq.~\ref{eqn:ViTstep5} and Fig.~\ref{fig:math}, we could give a more detailed explanation of the rationalization of robust bias. Inspired by~\cite{yin2019fourier, xie2020smooth,gowal2019scalable}, we know that adversarial training leads to the enhancement of model robust through learning adversarial examples generated from the inner high-frequency visual structures. Additionally, the inner maximization process could find more challenging adversarial examples by increasing the proportion of high-frequency structure, and the convolution operation would be a kind of good high-frequency visual structure. Therefore, we further name the convolution operation as robust bias since it could influence the adversarial robustness after adversarial training by modifying its proportion in a whole structure. Apart from using the experimental evaluation to certify the effect of robust bias, we also adopt a simple mathematical analysis here.
Fig.~\ref{fig:math} is the activation function (Sigmoid as Eq.~\ref{eq:sigmoid}) and loss function (Cross-entropy loss as Eq.~\ref{eq:cross-entropy}) for the most straightforward two-class classification problem. Since the Sigmod function is monotonically increasing, when the input of Sigmoid moves to frequency values from $X_1$ to $X_2$, the output will also change from $y_1$ to $y_2$. In the cross-entropy loss of two labels, the possible value range of loss would also extend from $l_1$ to $l_2$. Consequentially, when the input frequency $X$ is more toward the high-value region, this simple classification task will result in a broader range of possible loss values like $l_1$ to $l_2$. Furthermore, since the inner max in adversarial training is pursuing higher loss value within $\ell_p$-ball constraint, higher-frequency information exploration will finally lead to more challenging adversarial examples. Eventually, these harder adversarial examples would facilitate the process of adversarial training and make our proposed robust bias effectively increase the final structure robustness.

\begin{equation}\label{eq:sigmoid}
    y(x) = Sigmoid(x) = \frac{2}{1+e^{-x}}
\end{equation}

\begin{equation}\label{eq:cross-entropy}
    L(x) = - [ylog(\hat{y}) + (1-y)log(1-\hat{y})]
\end{equation}

\begin{table}[h!]
\small
\centering
\resizebox{\linewidth}{!}{ 
\begin{tabular}{cccccc|c|c|c}
\hline
\multicolumn{6}{c|}{ViT/VMLP}                                                                                                                                                                                                                                           & \multirow{2}{*}{\begin{tabular}[c]{@{}c@{}}Clean\\ Accuracy\end{tabular}} & \multirow{2}{*}{\begin{tabular}[c]{@{}c@{}}PGD\\ (4/255)\end{tabular}} & \multirow{2}{*}{\begin{tabular}[c]{@{}c@{}}Auto-\\ Attack\\ (4/255)\end{tabular}} \\ \cline{1-6}
\multicolumn{1}{c|}{\begin{tabular}[c]{@{}c@{}}Components\\ Combine\end{tabular}} & \multicolumn{1}{c|}{Embedding} & \multicolumn{1}{c|}{TM}     & \multicolumn{1}{c|}{CMLP}  & \multicolumn{1}{c|}{Norm} & \begin{tabular}[c]{@{}c@{}}Stacking\\ Structure\end{tabular} &                                                                           &                                                                        &                                                                         \\ \hline
\multicolumn{1}{c|}{(b)-Ori}                                                          & \multicolumn{1}{c|}{Ori}       & \multicolumn{1}{c|}{Ori}    & \multicolumn{1}{c|}{Ori}  & \multicolumn{1}{c|}{LN}   & oriViT                                                       & 58.63/57.93                                                               & 26.32/23.62                                                            & 25.98/22.79                                                             \\ 
\multicolumn{1}{c|}{(h)}                                                          & \multicolumn{1}{c|}{CONV}      & \multicolumn{1}{c|}{-}      & \multicolumn{1}{c|}{CONV} & \multicolumn{1}{c|}{-}    & -                                                            & 61.25/61.12                                                               & 28.58/27.43                                                            & 28.22/26.74                                                             \\ \hline
\multicolumn{1}{c|}{(i)-CVT}                                                          & \multicolumn{1}{c|}{PCONV}     & \multicolumn{1}{c|}{CONV}   & \multicolumn{1}{c|}{Ori}  & \multicolumn{1}{c|}{-}    & CNN-based                                                    & 57.43/56.97                                                               & 27.40/26.89                                                            & 26.64/25.98                                                             \\ 
\multicolumn{1}{c|}{(j)}                                                          & \multicolumn{1}{c|}{-}         & \multicolumn{1}{c|}{-}      & \multicolumn{1}{c|}{CONV} & \multicolumn{1}{c|}{-}    & -                                                            & 61.07/60.61                                                               & 30.28/29.99                                                            & 28.77/26.91                                                             \\ \hline
\multicolumn{1}{c|}{(k)-Swin}                                                          & \multicolumn{1}{c|}{Ori}       & \multicolumn{1}{c|}{WBM+SWBM} & \multicolumn{1}{c|}{Ori}  & \multicolumn{1}{c|}{-}    & Swin-based                                                   & 59.74/58.98                                                               & 24.12/21.62                                                            & 23.58/20.19                                                             \\ 
\multicolumn{1}{c|}{(l)}                                                          & \multicolumn{1}{c|}{PCONV}     & \multicolumn{1}{c|}{-}      & \multicolumn{1}{c|}{CONV} & \multicolumn{1}{c|}{-}    & -                                                            & 62.37/61.45                                                               & 28.13/27.24                                                            & 27.62/25.83                                                             \\ \hline
\multicolumn{1}{c|}{(m)-NT}                                                          & \multicolumn{1}{c|}{Ori}       & \multicolumn{1}{c|}{Ori}    & \multicolumn{1}{c|}{Ori}  & \multicolumn{1}{c|}{-}    & ImagePy                                                      & 58.37/57.98                                                               & 27.89/26.59                                                            & 26.67/25.76                                                             \\ 
\multicolumn{1}{c|}{(n)-RB}                                                          & \multicolumn{1}{c|}{PCONV}     & \multicolumn{1}{c|}{CONV}   & \multicolumn{1}{c|}{CONV} & \multicolumn{1}{c|}{-}    & -                                                            & \textbf{61.59/60.27}                                                      & \textbf{32.71/32.09}                                                   & \textbf{32.25/31.78}                                                    \\ \hline
\end{tabular}
}
\caption{\footnotesize{The robust performance of our proposed representative structures in ImageNet-1k.
All results are shown in \textbf{ViT accuracy (\%)/VMLP accuracy  (\%)}.
 (b) is the original ViT/VMLP~\cite{dosovitskiy2020image,tolstikhin2021mlp} (Ori), 
 (h) is the oriViT structure with the most convolution operation,
 (i) is corresponding to CVT~\cite{wu2021cvt}/CVT-based VMLP (CVT) or CNN-based structures, 
  (j) is the CNN-based structure with the most convolution operation,
 (k) is Swin ViT/MLP~\cite{liu2021swin} (Swin),  
 (l) is the Swin-based structure with the most convolution operation,
 (m) is the NesT~\cite{zhang2021aggregating}/NesT-based VMLP (NT), and (n) is our final RBViT/RBMLP (RB).
}}
\label{tab:imagenet}
\vspace{-5mm}
\end{table}

\begin{table}[h!]
\setlength{\tabcolsep}{4pt}
\centering
\adjustbox{max width=1.0\textwidth}{
\begin{tabular}{cccccc|c|ccc|ccc}
\hline
\multicolumn{6}{c|}{ViT/VMLP}                                                                                                                                                                                                                                                                                                                   & \multirow{2}{*}{\begin{tabular}[c]{@{}c@{}}Clean\\ Accuracy\end{tabular}} & \multicolumn{3}{c|}{PGD}                                                        & \multicolumn{3}{c}{\begin{tabular}[c]{@{}c@{}}Auto-\\ Attack\end{tabular}}        \\ \cline{1-6} \cline{8-13} 
\multicolumn{1}{c|}{\begin{tabular}[c]{@{}c@{}}Elem-\\ ents\end{tabular}} & \multicolumn{1}{c|}{\begin{tabular}[c]{@{}c@{}}Emb-\\ edding\end{tabular}} & \multicolumn{1}{c|}{\begin{tabular}[c]{@{}c@{}}TM\end{tabular}} & \multicolumn{1}{c|}{CMLP}  & \multicolumn{1}{c|}{Norm} & \begin{tabular}[c]{@{}c@{}}Skip-\\ Connect\end{tabular} &                                                                           & \multicolumn{1}{c|}{1/255}     & \multicolumn{1}{c|}{2/255}       & 3/255       & \multicolumn{1}{c|}{1/255}       & \multicolumn{1}{c|}{2/255}       & 3/255       \\ \hline
\multicolumn{1}{c|}{(1)}                                                  & \multicolumn{1}{c|}{Ori}                                                   & \multicolumn{1}{c|}{Ori}                                               & \multicolumn{1}{c|}{Ori}  & \multicolumn{1}{c|}{LN}   & Res                                                     & 85.7/81.4                                                                 & \multicolumn{1}{c|}{57.1/40.0} & \multicolumn{1}{c|}{52.70/44.06} & 52.70/44.06 & \multicolumn{1}{c|}{52.70/44.06} & \multicolumn{1}{c|}{52.70/44.06} & 52.70/44.06 \\ \hline
\multicolumn{1}{c|}{(2)}                                                  & \multicolumn{1}{c|}{-}                                                   & \multicolumn{1}{c|}{-}                                               & \multicolumn{1}{c|}{-}  & \multicolumn{1}{c|}{-}   & NONE                                                   & 72.5/80.5                                                                 & \multicolumn{1}{c|}{36.9/40.3} & \multicolumn{1}{c|}{11.6/19.0}   & 2.1/7.4     & \multicolumn{1}{c|}{36.7/39.6}   & \multicolumn{1}{c|}{10.5/17.6}   & 1.5/6.5     \\ \hline
\multicolumn{1}{c|}{(3)}                                                  & \multicolumn{1}{c|}{-}                                                   & \multicolumn{1}{c|}{-}                                               & \multicolumn{1}{c|}{-}  & \multicolumn{1}{c|}{NONE} & Res                                                     & 84.3/82.1                                                                 & \multicolumn{1}{c|}{57.3/44.9} & \multicolumn{1}{c|}{24.5/16.7}   & 8.2/4.9     & \multicolumn{1}{c|}{54.7/43.5}   & \multicolumn{1}{c|}{23.4/15.4}   & 6.8         \\ \hline
\multicolumn{1}{c|}{(4)}                                                  & \multicolumn{1}{c|}{-}                                                   & \multicolumn{1}{c|}{-}                                               & \multicolumn{1}{c|}{-}  & \multicolumn{1}{c|}{-} & NONE                                                    & 10/10                                                                     & \multicolumn{1}{c|}{10/10}     & \multicolumn{1}{c|}{10/10}       & 10/10       & \multicolumn{1}{c|}{10/10}       & \multicolumn{1}{c|}{10/10}       & 10/10       \\ \hline
\multicolumn{1}{c|}{(5)}                                                  & \multicolumn{1}{c|}{-}                                                   & \multicolumn{1}{c|}{-}                                               & \multicolumn{1}{c|}{NONE} & \multicolumn{1}{c|}{LN}   & Res                                                     & 78.2/61.4                                                                 & \multicolumn{1}{c|}{40.6/39.7} & \multicolumn{1}{c|}{17.8/20.3}   & 4.8/8.4     & \multicolumn{1}{c|}{37.8/38.5}   & \multicolumn{1}{c|}{16.5/18.7}   & 3.2/6.9     \\ \hline
\multicolumn{1}{c|}{(6)}                                                  & \multicolumn{1}{c|}{-}                                                   & \multicolumn{1}{c|}{-}                                               & \multicolumn{1}{c|}{-} & \multicolumn{1}{c|}{-}   & NONE                                                    & 70.34/59.4                                                                & \multicolumn{1}{c|}{33.4/39.2} & \multicolumn{1}{c|}{20.3/21.6}   & 5.1/9.9     & \multicolumn{1}{c|}{32.3/38.2}   & \multicolumn{1}{c|}{16.4/20.2}   & 7.1/9       \\ \hline
\multicolumn{1}{c|}{(7)}                                                  & \multicolumn{1}{c|}{-}                                                   & \multicolumn{1}{c|}{-}                                               & \multicolumn{1}{c|}{-} & \multicolumn{1}{c|}{NONE} & Res                                                     & 79.47/61.7                                                                & \multicolumn{1}{c|}{46.8/43.8} & \multicolumn{1}{c|}{27.4/25.8}   & 15.7/13     & \multicolumn{1}{c|}{45.9/42.2}   & \multicolumn{1}{c|}{23.5/21.7}   & 10.1/7.2    \\ \hline
\multicolumn{1}{c|}{(8)}                                                  & \multicolumn{1}{c|}{-}                                                   & \multicolumn{1}{c|}{-}                                               & \multicolumn{1}{c|}{-} & \multicolumn{1}{c|}{-} & NONE                                                    & 45.38/58.9                                                                & \multicolumn{1}{c|}{31.2/41.9} & \multicolumn{1}{c|}{14.5/25.7}   & 6.9/13.5    & \multicolumn{1}{c|}{34.1/39.5}   & \multicolumn{1}{c|}{22.5/21.9}   & 9.6/8.7     \\ \hline
\multicolumn{1}{c|}{(9)}                                                  & \multicolumn{1}{c|}{-}                                                   & \multicolumn{1}{c|}{-}                                               & \multicolumn{1}{c|}{CONV} & \multicolumn{1}{c|}{LN}   & Res                                                     & 87.4/85.0                                                                 & \multicolumn{1}{c|}{59.8/55.2} & \multicolumn{1}{c|}{27.7/26.2}   & 10.5/9.5    & \multicolumn{1}{c|}{57.8/53.7}   & \multicolumn{1}{c|}{25.4/24.6}   & 8.9/7.9     \\ \hline
\multicolumn{1}{c|}{(10)}                                                 & \multicolumn{1}{c|}{-}                                                   & \multicolumn{1}{c|}{-}                                               & \multicolumn{1}{c|}{-} & \multicolumn{1}{c|}{-}   & NONE                                                    & 79.36/10                                                                  & \multicolumn{1}{c|}{41.6/10}   & \multicolumn{1}{c|}{22.3/10}     & 9.8/10      & \multicolumn{1}{c|}{53.4/51.7}   & \multicolumn{1}{c|}{23.7/21.2}   & 8.9/10.2    \\ \hline
\multicolumn{1}{c|}{(11)}                                                 & \multicolumn{1}{c|}{-}                                                   & \multicolumn{1}{c|}{-}                                               & \multicolumn{1}{c|}{-} & \multicolumn{1}{c|}{NONE} & Res                                                     & 87.4/85.2                                                                 & \multicolumn{1}{c|}{59.8/58.9} & \multicolumn{1}{c|}{32.9/31.2}   & 11.2/9.5    & \multicolumn{1}{c|}{53.4/51.7}   & \multicolumn{1}{c|}{22.8/21.2}   & 9.2/6.4     \\ \hline
\multicolumn{1}{c|}{(12)}                                                 & \multicolumn{1}{c|}{-}                                                   & \multicolumn{1}{c|}{-}                                               & \multicolumn{1}{c|}{-} & \multicolumn{1}{c|}{-} & NONE                                                    & 10/10                                                                     & \multicolumn{1}{c|}{10/10}     & \multicolumn{1}{c|}{10/10}       & 10/10       & \multicolumn{1}{c|}{10/10}       & \multicolumn{1}{c|}{10/10}       & 10/10       \\ \hline
\multicolumn{1}{c|}{(13)}                                                 & \multicolumn{1}{c|}{CONV}                                                  & \multicolumn{1}{c|}{-}                                               & \multicolumn{1}{c|}{Ori}  & \multicolumn{1}{c|}{LN}   & Res                                                     & 88.1/84.6                                                                 & \multicolumn{1}{c|}{46.4/41.9} & \multicolumn{1}{c|}{13.8/10.9}   & 3.0/2.0     & \multicolumn{1}{c|}{42.4/40.1}   & \multicolumn{1}{c|}{11.3/9.1}    & 2.4/1.7     \\ \hline
\multicolumn{1}{c|}{(14)}                                                 & \multicolumn{1}{c|}{-}                                                  & \multicolumn{1}{c|}{-}                                               & \multicolumn{1}{c|}{-}  & \multicolumn{1}{c|}{-}   & NONE                                                    & 75.8/72.8                                                                 & \multicolumn{1}{c|}{19.8/23.1} & \multicolumn{1}{c|}{2.2/4.0}     & 0.2/0.5     & \multicolumn{1}{c|}{18.6/20.2}   & \multicolumn{1}{c|}{1.5/2.3}     & 0/0         \\ \hline
\multicolumn{1}{c|}{(15)}                                                 & \multicolumn{1}{c|}{-}                                                  & \multicolumn{1}{c|}{-}                                               & \multicolumn{1}{c|}{-}  & \multicolumn{1}{c|}{NONE} & Res                                                     & 89.2/84.7                                                                 & \multicolumn{1}{c|}{56.6/50.7} & \multicolumn{1}{c|}{21.2/17.9}   & 4.9/4.2     & \multicolumn{1}{c|}{55.3/38.2}   & \multicolumn{1}{c|}{17.1/16.4}   & 2.7/2.1     \\ \hline
\multicolumn{1}{c|}{(16)}                                                 & \multicolumn{1}{c|}{-}                                                  & \multicolumn{1}{c|}{-}                                               & \multicolumn{1}{c|}{-}  & \multicolumn{1}{c|}{-} & NONE                                                    & 10/10                                                                     & \multicolumn{1}{c|}{10/10}     & \multicolumn{1}{c|}{10/10}       & 10/10       & \multicolumn{1}{c|}{10/10}       & \multicolumn{1}{c|}{10/10}       & 10/10       \\ \hline
\multicolumn{1}{c|}{(17)}                                                 & \multicolumn{1}{c|}{-}                                                  & \multicolumn{1}{c|}{-}                                               & \multicolumn{1}{c|}{NONE} & \multicolumn{1}{c|}{LN}   & Res                                                     & 79.0/68.7                                                                 & \multicolumn{1}{c|}{30.6/20.5} & \multicolumn{1}{c|}{5.4/3.1}     & 0.6/0.2     & \multicolumn{1}{c|}{28.7/22.4}   & \multicolumn{1}{c|}{6.7/5.4}     & 1.6/1.2     \\ \hline
\multicolumn{1}{c|}{(18)}                                                 & \multicolumn{1}{c|}{-}                                                  & \multicolumn{1}{c|}{-}                                               & \multicolumn{1}{c|}{-} & \multicolumn{1}{c|}{-}   & NONE                                                    & 73.8/67.4                                                                 & \multicolumn{1}{c|}{17.1/21.2} & \multicolumn{1}{c|}{1.0/3.2}     & 0.6/0.3     & \multicolumn{1}{c|}{15.4/19.2}   & \multicolumn{1}{c|}{1.2/0.6}     & 0/0         \\ \hline
\multicolumn{1}{c|}{(19)}                                                 & \multicolumn{1}{c|}{-}                                                  & \multicolumn{1}{c|}{-}                                               & \multicolumn{1}{c|}{-} & \multicolumn{1}{c|}{NONE} & Res                                                     & 80.9/68.4                                                                 & \multicolumn{1}{c|}{45/38.4}   & \multicolumn{1}{c|}{14.5/14.0}   & 2.6/3.4     & \multicolumn{1}{c|}{43.1/36.7}   & \multicolumn{1}{c|}{13.2/12.8}   & 2.1/3.2     \\ \hline
\multicolumn{1}{c|}{(20)}                                                 & \multicolumn{1}{c|}{-}                                                  & \multicolumn{1}{c|}{-}                                               & \multicolumn{1}{c|}{-} & \multicolumn{1}{c|}{-} & NONE                                                    & 76.0/68.4                                                                 & \multicolumn{1}{c|}{40.8/35.8} & \multicolumn{1}{c|}{11.8/12.5}   & 1.9/2.7     & \multicolumn{1}{c|}{38.4/34.2}   & \multicolumn{1}{c|}{8.4/7.9}     & 1.2/2.4     \\ \hline
\multicolumn{1}{c|}{(21)}                                                 & \multicolumn{1}{c|}{-}                                                  & \multicolumn{1}{c|}{-}                                               & \multicolumn{1}{c|}{CONV} & \multicolumn{1}{c|}{LN}   & Res                                                     & 89.8/88.1                                                                 & \multicolumn{1}{c|}{59.2/59.9} & \multicolumn{1}{c|}{24.0/27.9}   & 6.8/9.6     & \multicolumn{1}{c|}{58.7/58.4}   & \multicolumn{1}{c|}{22.4/21.9}   & 5.9/6.0     \\ \hline
\multicolumn{1}{c|}{(22)}                                                 & \multicolumn{1}{c|}{-}                                                  & \multicolumn{1}{c|}{-}                                               & \multicolumn{1}{c|}{-} & \multicolumn{1}{c|}{-}   & NONE                                                    & 82.6/10                                                                   & \multicolumn{1}{c|}{34.6/10}   & \multicolumn{1}{c|}{10.2/6.3}    & 4.3/0       & \multicolumn{1}{c|}{33.4/10}     & \multicolumn{1}{c|}{10.0/5.4}    & 3.3/0       \\ \hline
\multicolumn{1}{c|}{(23)}                                                 & \multicolumn{1}{c|}{-}                                                  & \multicolumn{1}{c|}{-}                                               & \multicolumn{1}{c|}{-} & \multicolumn{1}{c|}{NONE} & Res                                                     & 89.7/88.1                                                                 & \multicolumn{1}{c|}{58.2/49.9} & \multicolumn{1}{c|}{17.5/17.2}   & 4.3/5.2     & \multicolumn{1}{c|}{57.2/48.3}   & \multicolumn{1}{c|}{16.3/15.9}   & 4.1/3.9     \\ \hline
\multicolumn{1}{c|}{(24)}                                                 & \multicolumn{1}{c|}{-}                                                  & \multicolumn{1}{c|}{-}                                               & \multicolumn{1}{c|}{-} & \multicolumn{1}{c|}{-} & NONE                                                    & 10/10                                                                     & \multicolumn{1}{c|}{10/10}     & \multicolumn{1}{c|}{10/10}       & 10/10       & \multicolumn{1}{c|}{10/10}       & \multicolumn{1}{c|}{10/10}       & 10/10       \\ \hline
\end{tabular}
}
\caption{\footnotesize{Results of the proposed 24 naturally trained structures. All results are shown in \textbf{ViT accuracy (\%)/VMLP accuracy (\%)}.
 All structures in this table only do not refer to various multi-hierarchy layer stacking and all keep to be OriViT. 
 The specific elements include (1) Embedding (ori/CONV); (2) TM block (Ori); (3) CMLP block (Ori/CONV/NONE); (4) Norm (LN/NONE), and 5)Skip-Connection (Res/NONE).}  } 
\label{tab:natural-alltrain}
\end{table}

\begin{table}[h]
\setlength{\tabcolsep}{4pt}
\centering
  \adjustbox{max width=1.0\textwidth}{
  \begin{tabular}{cccccc|c|c|c|c}
\toprule
\multicolumn{6}{c|}{ViT/VMLP}                         & \multirow{2}{*}{\begin{tabular}[c]{@{}c@{}}Clean\\ Accuracy\end{tabular}}
& 
\multicolumn{1}{c|}{\multirow{2}{*}{\begin{tabular}[c]{@{}c@{}}PGD\\ (8/255)\end{tabular}}} 
& 
\multirow{2}{*}{\begin{tabular}[c]{@{}c@{}}Auto-\\ Attack\\ (8/255)\end{tabular}} 
& 
\multirow{2}{*}{\begin{tabular}[c]{@{}c@{}}Lipschitz\\ Constant\end{tabular}} \\ \cline{1-6}
\multicolumn{1}{c|}{\begin{tabular}[c]{@{}c@{}}Stru-\\ ctures\end{tabular}} & \multicolumn{1}{c|}{\begin{tabular}[c]{@{}c@{}}Emb-\\ edding\end{tabular}} & \multicolumn{1}{c|}{\begin{tabular}[c]{@{}c@{}}TM\end{tabular}} & \multicolumn{1}{c|}{CMLP}  & \multicolumn{1}{c|}{Norm} & \begin{tabular}[c]{@{}c@{}}Skip-\\ Connect\end{tabular} &                                                                           & \multicolumn{1}{l|}{}                                                                       &                                                                                   &                                                                               \\ \midrule
\multicolumn{1}{c|}{(1)}                                                  & \multicolumn{1}{c|}{Ori}                                                   & \multicolumn{1}{c|}{Ori}                                               & \multicolumn{1}{c|}{Ori}  & \multicolumn{1}{c|}{LN}   & Res                                                     & 79.93/66.38                                                               & 52.70/44.06                                                              &  51.45/43.10                                                                                 &   157.7/164.7                                                                            \\ \hline
\multicolumn{1}{c|}{(2)}                                                  & \multicolumn{1}{c|}{-}                                                   & \multicolumn{1}{c|}{-}                                               & \multicolumn{1}{c|}{-}  & \multicolumn{1}{c|}{-}   & NONE                                                    & 52.59/53.14                                                               & 37.10/37.63                                                          &     35.89/35.66                                                                &        169.3/168.2                                                                        \\ \hline
\multicolumn{1}{c|}{(3)}                                                  & \multicolumn{1}{c|}{-}                                                   & \multicolumn{1}{c|}{-}                                               & \multicolumn{1}{c|}{-}  & \multicolumn{1}{c|}{None} & Res    & 79.88/71.06              & 52.66/45.56         & 51.12/44.37    &      159.2/163.2                                                                         \\ \hline
\multicolumn{1}{c|}{(4)}                                                  & \multicolumn{1}{c|}{-}                                                   & \multicolumn{1}{c|}{-}                                               & \multicolumn{1}{c|}{-}  & \multicolumn{1}{c|}{-} & NONE                                                    & 10/10                                                                     & 10/10                                                                   &   10/10                                                                     &  0/0                                                                          \\ \hline
\multicolumn{1}{c|}{(5)}                                                  & \multicolumn{1}{c|}{-}                                                   & \multicolumn{1}{c|}{-}                                               & \multicolumn{1}{c|}{NONE} & \multicolumn{1}{c|}{LN}   & Res                                                     & 55.80/49.22                                                               & 38.69/36.35                                                                  &    37.45/34.98                                                     &    167.4/172.3                                                                           \\ \hline
\multicolumn{1}{c|}{(6)}                                                  & \multicolumn{1}{c|}{-}                                                   & \multicolumn{1}{c|}{-}                                               & \multicolumn{1}{c|}{-} & \multicolumn{1}{c|}{-}   & NONE                                                    & 56.45/50.34                                                               & 39.26/36.15                                                     &     39.56/34.96                                              &         165.1/173.9                                                                      \\ \hline
\multicolumn{1}{c|}{(7)}                                                  & \multicolumn{1}{c|}{-}                                                   & \multicolumn{1}{c|}{-}                                               & \multicolumn{1}{c|}{-} & \multicolumn{1}{c|}{NONE} & Res                                                     & 58.88/49.89                                                               & 42.25/36.35                                                          &    41.76/35.89                                &     152.4/173.3                                                                          \\ \hline
\multicolumn{1}{c|}{(8)}                                                  & \multicolumn{1}{c|}{-}                                                   & \multicolumn{1}{c|}{-}                                               & \multicolumn{1}{c|}{-} & \multicolumn{1}{c|}{-} & NONE                                                    & 46.79/48.70                                                               & 34.99/36.71                                                          &     34.14/35.34                                 &         180.6/170.3                                                                      \\ \hline
\multicolumn{1}{c|}{(9)}                                                  & \multicolumn{1}{c|}{-}                                                   & \multicolumn{1}{c|}{-}                                               & \multicolumn{1}{c|}{CONV} & \multicolumn{1}{c|}{LN}   & Res                                                     & 81.66/77.58                                                               & 54.69/51.00                                      &      53.85/50.88                                                   &   152.7/157.5                                                                            \\ \hline
\multicolumn{1}{c|}{(10)}                                                 & \multicolumn{1}{c|}{-}                                                   & \multicolumn{1}{c|}{-}                                               & \multicolumn{1}{c|}{-} & \multicolumn{1}{c|}{-}   & NONE                                                    & 10/10                                                                     & 10/10                                                       &     10/10                                                                              & 0/0                                                                              \\ \hline
\multicolumn{1}{c|}{(11)}                                                 & \multicolumn{1}{c|}{-}                                                   & \multicolumn{1}{c|}{-}                                               & \multicolumn{1}{c|}{-} & \multicolumn{1}{c|}{NONE} & Res                                                     & 82.81/78.83                                                               & 54.79/54.24                                                           &   53.83/53.69                                                       &  151.3/152.3                                                                             \\ \hline
\multicolumn{1}{c|}{(12)}                                                 & \multicolumn{1}{c|}{-}                                                   & \multicolumn{1}{c|}{-}                                               & \multicolumn{1}{c|}{-} & \multicolumn{1}{c|}{-} & NONE                                                    & 10/10                                                                     & 10/10                                                           &     10/10                                                   &     0/0                                                                          \\ \hline
\multicolumn{1}{c|}{(13)}                                                 & \multicolumn{1}{c|}{CONV}                                                  & \multicolumn{1}{c|}{-}                                               & \multicolumn{1}{c|}{Ori}  & \multicolumn{1}{c|}{LN}   & Res                                                     & 80.50/75.92                                                               & 54.40/50.89                                           &       53.69/48.99                                                    &   153.1/162.3                                                                            \\ \hline
\multicolumn{1}{c|}{(14)}                                                 & \multicolumn{1}{c|}{-}                                                  & \multicolumn{1}{c|}{-}                                               & \multicolumn{1}{c|}{-}  & \multicolumn{1}{c|}{-}   & NONE                                                    & 64.75/62.08                                                               & 44.49/42.14                                                  &      43.47/41.81                                              &    163.2/166.7                                                                           \\ \hline
\multicolumn{1}{c|}{(15)}                                                 & \multicolumn{1}{c|}{-}                                                  & \multicolumn{1}{c|}{-}                                               & \multicolumn{1}{c|}{-}  & \multicolumn{1}{c|}{NONE} & Res                                                     & 82.77/77.86                                                               & 55.85/53.22                                                   &       54.98/52.89                                             &      151.4/155.8                                                                         \\ \hline
\multicolumn{1}{c|}{(16)}                                                 & \multicolumn{1}{c|}{-}                                                  & \multicolumn{1}{c|}{-}                                               & \multicolumn{1}{c|}{-}  & \multicolumn{1}{c|}{-} & None                                                    & 10/10                                                                     & 10/10                                                        &      10/10                                                   &    0/0                                                                           \\ \hline
\multicolumn{1}{c|}{(17)}                                                 & \multicolumn{1}{c|}{-}                                                  & \multicolumn{1}{c|}{-}                                               & \multicolumn{1}{c|}{NONE} & \multicolumn{1}{c|}{LN}   & Res                                                     & 71.12/56.11                                                               & 48.35/41.13                                                       &     47.78/39.89                                            &       161.3/168.2                                                                        \\ \hline
\multicolumn{1}{c|}{(18)}                                                 & \multicolumn{1}{c|}{-}                                                  & \multicolumn{1}{c|}{-}                                               & \multicolumn{1}{c|}{-} & \multicolumn{1}{c|}{-}   & NONE                                                    & 60.01/54.39                                                               & 42.40/39.61                                                   &     41.56/38.44                                                 &     167.4/171.1                                                                          \\ \hline
\multicolumn{1}{c|}{(19)}                                                 & \multicolumn{1}{c|}{-}                                                  & \multicolumn{1}{c|}{-}                                               & \multicolumn{1}{c|}{-} & \multicolumn{1}{c|}{NONE} & Res                                                     & 75.01/56.25                                                               & 52.92/39.81                                         &        52.10/38.96                                                                         &   158.7/166.9                                                                              \\ \hline
\multicolumn{1}{c|}{(20)}                                                 & \multicolumn{1}{c|}{-}                                                  & \multicolumn{1}{c|}{-}                                               & \multicolumn{1}{c|}{-} & \multicolumn{1}{c|}{-} & NONE                                                    & 64.61/55.80                                                               & 46.80/41.31                                                   &     46.90/40.65                                                   &      163.4/161.5                                                                         \\ \hline
\multicolumn{1}{c|}{(21)}                                                 & \multicolumn{1}{c|}{-}                                                  & \multicolumn{1}{c|}{-}                                               & \multicolumn{1}{c|}{CONV} & \multicolumn{1}{c|}{LN}   & Res                                                     & \textbf{82.35/81.42 }                                                              & \textbf{56.41/56.89}                                                                                 &     \textbf{56.12/57.02}                                                                              &       \textbf{140.3/141.26   }                                                                     \\ \hline
\multicolumn{1}{c|}{(22)}                                                 & \multicolumn{1}{c|}{-}                                                  & \multicolumn{1}{c|}{-}                                               & \multicolumn{1}{c|}{-} & \multicolumn{1}{c|}{-}   & NONE                                                    & 15.53/10                                                                  & 10/10                                                                                       &  10/10                                                  &    0/0                                                                           \\ \hline
\multicolumn{1}{c|}{(23)}                                                 & \multicolumn{1}{c|}{-}                                                  & \multicolumn{1}{c|}{-}                                               & \multicolumn{1}{c|}{-} & \multicolumn{1}{c|}{NONE} & Res                                                     & 80.57/79.25                                                             & 55.63/53.81                                                             &    54.23/52.45                                                                  &       146.3/148.5                                                                       \\ \hline
\multicolumn{1}{c|}{(24)}                                                 & \multicolumn{1}{c|}{-}                                                  & \multicolumn{1}{c|}{-}                                               & \multicolumn{1}{c|}{-} & \multicolumn{1}{c|}{-} & NONE                                                    & 10/10                                                                     & 10/10                                                                                       &  10/10                                                      &   0/0                                                                            \\ \bottomrule
\end{tabular}
}
\caption{\footnotesize{Results of the proposed 24 adversarially trained structures in ViT/VMLP. All results are shown in \textbf{ViT accuracy (\%)/VMLP accuracy (\%)}. 
 All structures in this table only do not refer to various multi-hierarchy layer stacking, and all keep to be OriViT. 
 The specific elements include (1) Embedding (ori/CONV), (2) TM block (Ori), (3) CMLP block (Ori/CONV/NONE MLP); (4) Norm (LN/NONE), and (5)Skip-Connection (Res/NONE).}} 
 \label{tab:Adv-alltrainitem}
\end{table}

\end{document}